\documentclass[runningheads]{llncs}

\usepackage{graphicx}
\usepackage[main=american]{babel} 
\usepackage{amsmath,amssymb} 
\usepackage{microtype} 
\usepackage[babel,autostyle]{csquotes} 

\usepackage[lined,boxed]{algorithm2e}
\newcommand{\ssl}{\textsc{ssl}}
\newcommand{\tsvm}{\textsc{tsvm}}
\newcommand{\svm}{\textsc{svm}}
\newcommand{\cccp}{\textsc{cccp}}
\newcommand{\mse}{\textsc{mse}}

\begin{document}

\title{Deep Low-Density Separation for Semi-Supervised Classification%
	\thanks{Work completed while K.S. interned at Adobe.}}
\titlerunning{Deep Low-Density Separation}
\author{Michael C. Burkhart\inst{1} \and
	Kyle Shan\inst{2}}
\authorrunning{M.C. Burkhart and K. Shan}
\institute{Adobe Inc., San Jos\'e, USA\\
	\email{mburkhar@adobe.com} \\ \and
	Stanford University, Stanford, USA\\
	\email{kylecshan@gmail.com}}
\maketitle

\begin{abstract}
	Given a small set of labeled data and a large set of unlabeled data,
	semi-supervised learning (\ssl{}) attempts to leverage the location of the
	unlabeled datapoints in order to create a better classifier than could be
	obtained from supervised methods applied to the labeled training set alone.
	Effective \ssl{} imposes structural assumptions on the data, e.g. that
	neighbors are more likely to share a classification or that the decision
	boundary lies in an area of low density. For complex and high-dimensional
	data, neural networks can learn feature embeddings to which traditional
	\ssl{} methods can then be applied in what we call hybrid methods.

	Previously-developed hybrid methods iterate between refining a latent
	representation and performing graph-based \ssl{} on this representation. In
	this paper, we introduce a novel hybrid method that instead applies
	low-density separation to the embedded features. We describe it in detail
	and discuss why low-density separation may be better suited for \ssl{} on
	neural network-based embeddings than graph-based algorithms. We validate
	our method using in-house customer survey data and compare it to other
	state-of-the-art learning methods. Our approach effectively classifies
	thousands of unlabeled users from a relatively small number of
	hand-classified examples.
\end{abstract}

	

\section{Background}

In this section, we describe the problem of semi-supervised learning (\ssl{})
from a mathematical perspective. We then outline some of the current approaches
to solve this problem, emphasizing those relevant to our current work.

\subsection{Problem Description}
Consider a small labeled training set $\mathcal D_{0} = \{(x_1,y_1), (x_2,y_2),
	\dotsc, (x_\ell, y_\ell)\}$ of vector-valued features $x_i \in\bbbr^d$ and
discrete-valued labels $y_i \in \{1,\dotsc, c\}$, for $1\leq i \leq \ell$.
Suppose we have a large set $\mathcal{D}_1 = \{x_{\ell+1}, x_{\ell+2},\dotsc,
	x_{\ell+u}\}$ of unlabeled features to which we would like to assign labels.
One could perform supervised learning on the labeled dataset $\mathcal{D}_0$ to
obtain a general classifier and then apply this classifier to $\mathcal{D}_1$.
However, this approach ignores any information about the distribution of the
feature-points contained in $\mathcal{D}_1$. In contrast, \ssl{} attempts to
leverage this additional information in order to either inductively train a
generalized classifier on the feature space or transductively assign labels
only to the feature-points in $\mathcal{D}_1$.

\begin{figure}[h]
	\centering
	\begin{minipage}[t]{0.45\textwidth}
		\includegraphics[width=0.95\linewidth]{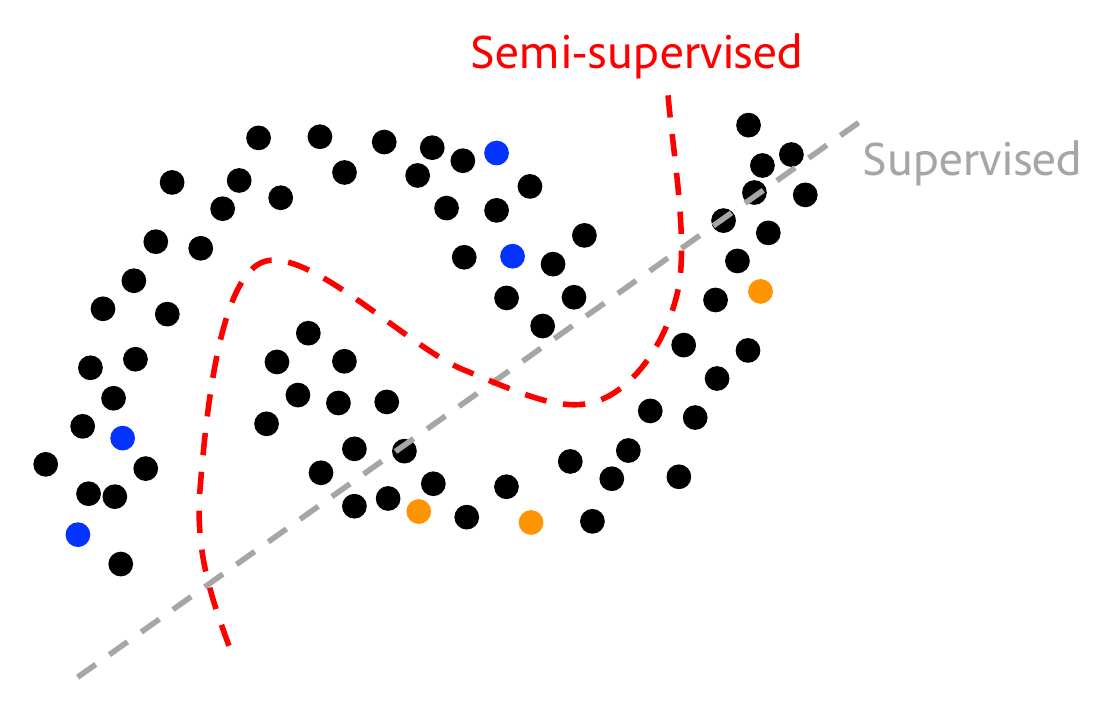}
		\caption{A schematic for semi-supervised classification. The grey line
			corresponds to a decision boundary obtained from a generic supervised
			classifier (incorporating information only from the labeled blue and
			orange points); the red line corresponds to a boundary from a generic
			semi-supervised method seeking a low-density decision boundary.}
		\label{fig:toy_example}
	\end{minipage}%
	\hfill
	\begin{minipage}[t]{0.45\textwidth}
		\includegraphics[width=0.95\linewidth]{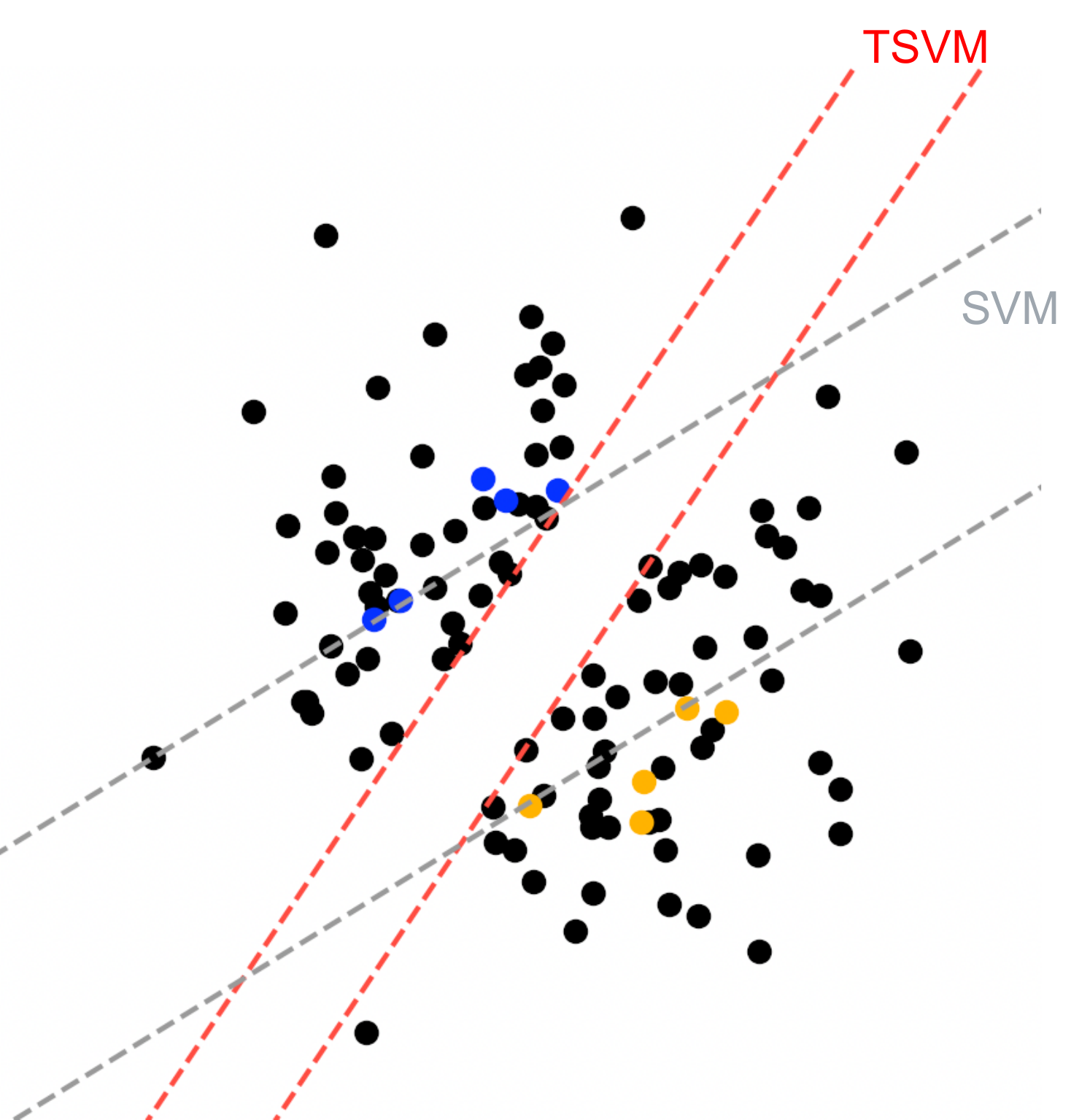}
		\caption{A schematic for \tsvm{} segmentation. The grey lines
			correspond to maximum margin separation for labeled data using a
			standard \svm{}; the red lines correspond to additionally penalizing
			unlabeled points that lie in the margin. In this example, the data is
			perfectly separable in two dimensions, but this need not always be
			true.}
		\label{fig:tsvm_diagram}
	\end{minipage}
\end{figure}

Effective \ssl{} methods impose additional assumptions about the structure of
the feature-data (i.e., $\{x: (x,y) \in \mathcal{D}_0\} \cup \mathcal{D}_1$);
for example, that features sharing the same label are clustered, that the
decision boundary separating differently labeled features is smooth, or that
the features lie on a lower dimensional manifold within $\bbbr^d$. In practice,
semi-supervised methods that leverage data from $\mathcal{D}_1$ can achieve
much better performance than supervised methods that use $\mathcal{D}_0$ alone.
See Figure~\ref{fig:toy_example} for a visualization. We describe both
graph-based and low-density separation methods along with neural network-based
approaches, as these are most closely related to our work. For a full survey,
see~\cite{Cha06,See01,Zhu05}.

\subsection{Graph-Based Methods}
Graph-based methods calculate the pairwise similarities between labeled and
unlabeled feature-points and allow labeled feature-points to pass labels to
their unlabeled neighbors. For example, label propagation~\cite{Zhu02} forms a
$(\ell+u)\times(\ell+u)$ dimensional transition matrix $T$ with transition
probabilities proportional to similarities (kernelized distances) between
feature-points and an $(\ell+u)\times c$ dimensional matrix of class
probabilities, and (after potentially smoothing this matrix) iteratively sets
$Y \leftarrow TY$, row-normalizes the probability vectors, and resets the rows
of probability vectors corresponding to the already-labeled elements of
$\mathcal{D}_0$. Label spreading~\cite{Zho04} follows a similar approach but
normalizes its weight matrix and allows for a (typically hand-tuned) clamping
parameter that assigns a level of uncertainty to the labels in $\mathcal{D}_0$.
There are many variations to the graph-based approach, including those that use
graph min-cuts~\cite{Blu01} and Markov random walks~\cite{Szu02}.

\subsection{Low-Density Separation}\label{s:low_d_sep}
Low-density separation methods attempt to find a decision boundary that best
separates one class of labeled data from the other. The quintessential example
is the transductive support vector machine
(\tsvm{}:~\cite{Gam98,Ben98,Joa99,Cha05,Cha08,Li15}), a semi-supervised
maximum-margin classifier of which there have been numerous variations. As
compared to the standard \svm{} (\emph{cf.}, e.g.,~\cite{Bis06,Mur12}), the
\tsvm{} additionally penalizes unlabeled points that lie close to the decision
boundary. In particular, for a binary classification problem with labels $y_i
	\in \{-1, 1\}$, it seeks parameters $w,b$ that minimize the non-convex
objective function
\begin{equation} \label{eq:tsvm_loss}
	J(w, b) = \frac12 \|w\|^2 + C \sum_{i=1}^l H(y_i \cdot f_{w, b}(x_i))
	+ C^* \sum_{i=l+1}^u H(|f_{w,b}(x_i)|),
\end{equation}
where $f_{w,b}: \mathbb{R}^d \to \mathbb{R}$ is the linear decision function
$f_{w,b}(x) = w \cdot x + b$, and $H(x) = \max(0, 1-x)$ is the hinge loss
function. The hyperparameters $C$ and $C^*$ control the relative influence of
the labeled and unlabeled data, respectively. Note that the third term,
corresponding to a loss function for the unlabeled data, is non-convex,
providing a challenge to optimization. See Figure~\ref{fig:tsvm_diagram} for a
visualization of how the \tsvm{} is intended to work and Ding et al.
\cite{Din17} for a survey of semi-supervised \svm{}'s. Other methods for
low-density separation include the more general entropy minimization
approach~\cite{Gra04}, along with information regularization~\cite{Szu02b} and
a Gaussian process-based approach~\cite{Law05}.

\subsection{Neural Network-Based Embeddings}
Both the graph-based and low-density separation approaches to \ssl{} rely on
the geometry of the feature-space providing a reasonable approximation to the
true underlying characteristics of the users or objects of interest. As
datasets become increasingly complex and high-dimensional, Euclidean distance
between feature vectors may not prove to be the best proxy for user or item
similarity. As the Gaussian kernel is a monotonic function of Euclidean
distance, kernelized methods such as label propagation and label spreading also
suffer from this criticism. While kernel learning approaches pose one potential
solution~\cite{Cha03,Zhu05b}, neural network-based embeddings have become
increasingly popular in recent years. Variational autoencoders (\textsc{vae}'s:
\cite{Kin14b}) and generative adversarial nets
(\textsc{gan}'s:~\cite{Li17,Dai17}) have both been successfully used for
\ssl{}. However, optimizing the parameters for these types of networks can
require expert hand-tuning and/or prohibitive computational
expense~\cite{Zop17,Rea17}. Additionally, most research in the area
concentrates on computer vision problems, and it is not clear how readily the
architectures and techniques developed for image classification translate to
other domains of interest.

\subsection{Hybrid Methods}
Recently, Iscen et al. introduced a neural embedding-based method to generate
features on which to perform label propagation~\cite{Isc19}. They train a
neural network-based classifier on the supervised dataset and then embed all
feature-points into an intermediate representation space. They then iterate
between performing label propagation in this feature space and continuing to
train their neural network classifier using weighted predictions from label
propagation (see also~\cite{Zhu19}). As these procedures are similar in spirit
to ours, we next outline our method in the next section and provide more
details as part of a comparison in subsection~\ref{sec:hybrid_comparison}.

\section{Deep Low-Density Separation Algorithm} \label{sec:methods}

In this section, we provide a general overview of our algorithm for deep
low-density separation and then delve into some of the details. We characterize
our general process as follows:

\begin{enumerate}
	\item We first learn a neural network embedding $f: \bbbr^d \to \bbbr^m$
	      for our feature-data optimized to differentiate between class labels. We
	      define a network $g:\bbbr^m \to \bbbp^c$ (initialized as the initial layers
	      from an autoencoder for the feature-data), where $\bbbp^c$ is the space of
	      $c$-dimensional probability vectors, and optimize $g\circ f$ on our labeled
	      dataset $\mathcal{D}_0$, where we one-hot encode the categories
	      corresponding to each $y_i$.
	\item We map all of the feature-points through this deep embedding and then
	      implement one-vs.-rest \tsvm{}'s for each class on this embedded data to
	      learn class-propensities for each unlabeled data point. We augment our
	      training data with the $x_i$ from $\mathcal{D}_1$ paired with the
	      propensities returned by this method and continue to train $g\circ f$ on
	      this new dataset for a few epochs.
	\item Our neural network $f$ now provides an even better embedding for
	      differentiating between classes. We repeat step 2 for a few iterations in
	      order for the better embedding to improve \tsvm{} separation, which upon
	      further training yields an even better embedding, and so forth, etc.
\end{enumerate}

\begin{algorithm}[h]
	\SetAlgoLined
	\KwData{labeled dataset $\mathcal{D}_0$ and unlabeled dataset $\mathcal{D}_1$}
	\KwResult{probabilistic predictions for the labels in $\mathcal{D}_1$}
	Initialize a deep neural network $f_\theta: \bbbr^d\to\bbbr^m$ with
	trainable parameters $\theta$\; Initialize a neural network $g_\psi:
		\bbbr^m\to\bbbp^c$ with trainable parameters $\psi$\; Obtain $\theta_0,
		\psi_0$ by minimizing cross entropy between $h(y_i)$ and
	$g_\psi(f_\theta(x_i))$ for $(x_i,y_i)\in\mathcal{D}_0$, where $h$ is the
	encoding defined in \eqref{eq:h}\;
	\For{$t = 1,\dotsc, T$}{
		Compute $\tilde D_0 = \{ (f_{\theta_{t-1}}(x), y) : (x,y) \in
			\mathcal{D}_0\}$ and $\tilde D_1 = \{ f_{\theta_{t-1}}(x) : x \in
			\mathcal{D}_1\}$\;
		Perform one-vs.-rest \tsvm{} training on
		$\tilde{\mathcal{D}}_0$ and $\tilde{\mathcal{D}}_1$ to obtain predicted
		probabilities $\hat p_i, i = \ell +1,\dotsc, \ell + u$ that the $x_i$ in
		$\mathcal{D}_1$ lie in each class and then set $\breve{\mathcal D}_1 =
			\{(x_i, \hat p_i)\}$ \;
		Obtain $\theta_t, \psi_t$ by continuing to optimize
		$g_\psi\circ f_\theta$, using $\mathcal{D}_0\cup \breve{\mathcal D}_1$\;
	}
	\KwRet $g_{\psi_T}(f_{\theta_T}(x_i))$ or an exponential moving average of
	the probabilistic predictions $g_{\psi_t}(f_{\theta_t}(x_i))$ for $x_i \in
	\mathcal{D}_0$.
	\caption{the Deep Segmentation Algorithm}
	\label{algo}
\end{algorithm}

This is our basic methodology, summarized as pseudo-code in
Algorithm~\ref{algo}. Upon completion, it returns a neural network $g\circ f$
that maps feature-values to class/label propensities that can easily be applied
to $\mathcal{D}_1$ and solve our problem of interest. In practice, we find that
taking an exponentially decaying moving average of the returned probabilities
as the algorithm progresses provides a slightly improved estimate. At each
iteration of the algorithm, we reinitialize the labels for the unlabeled points
and allow the semi-supervised \tsvm{} to make inferences using the new
embedding of the feature-data alone. In this way, it is possible to recover
from mistakes in labeling that occurred in previous iterations of the
algorithm.

\subsection{Details: Neural Network Training}
In our instantiation, the neural network $f:\bbbr^d\to \bbbr^m$ has two layers,
the first of size 128 and the second of size 32, both with hyperbolic tangent
activation. In between these two layers, we apply batch
normalization~\cite{Iof15} followed by dropout at a rate of 0.5 during model
training to prevent overfitting~\cite{Sri14}. The neural network
$g:\bbbr^m\to\bbbp^c$ consists of a single layer with 5 units and softmax
activation. We let $\theta$ (resp. $\psi$) denote the trainable parameters for
$f$ (resp. $g$) and sometimes use the notation $f_\theta$ and $g_\psi$ to
stress the dependence of the neural networks on these trainable parameters.
Neural network parameters receive Glorot normal initialization~\cite{Glo10}.
The network weights for $f$ and $g$ receive
Tikhonov-regularization~\cite{Tik43,Tik63}, which decreases as one progresses
through the network.

We form our underlying target distribution by one-hot encoding the labels $y_i$
and slightly smoothing these labels. We define $h:\{1,\dotsc, c\} \to \bbbp^c$
by its components $1\leq j \leq c$ as
\begin{equation}\label{eq:h}
	h(y)_j = \begin{cases} 1-c\cdot\epsilon, & \text{if } y=j, \\ 
		\epsilon, & \text {otherwise} \end{cases}
\end{equation}
where we set $\epsilon=10^{-3}$ to be our smoothing parameter.

Training proceeds as follows. We begin by training the neural network
$f_\theta$ to minimize $D_{\textsc{kl}} \big(h(y_i) \vert\vert
	g_{\psi}(f_{\theta}(x_i))\big)$ the Kullback--Leibler (\textsc{kl}) divergence
between the true distributions $h(y_i)$ and our inferred distributions
$g_{\psi}(f_{\theta}(x_i))$, on $\mathcal{D}_0$ in batches. For parameter
updates, we use the Adam optimizer~\cite{Kin15} that maintains different
learning rates for each parameter like AdaGrad~\cite{Duc11} and allows these
rates to sometimes increase like Adadelta~\cite{Zei12} but adapts them based on
the first two moments from recent gradient updates. This optimization on
labeled data produces parameters $\theta_0$ for $f$ and $\psi_0$ for $g$.

\begin{figure*}[h]
	\begin{center}
		\includegraphics[width=\textwidth]{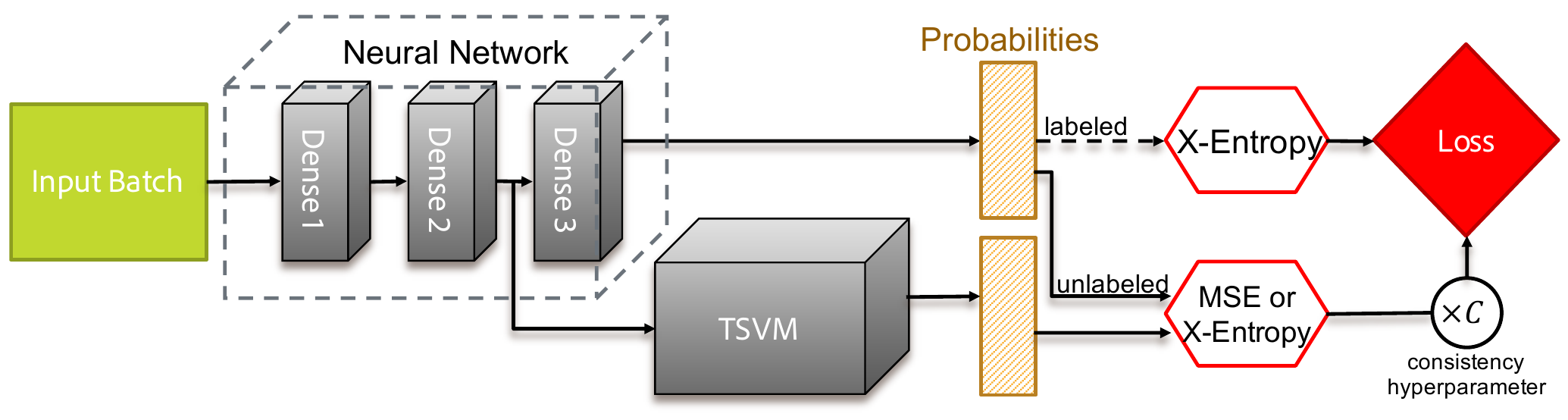}
		\caption{A schematic for Deep Low-Density Separation. The first two
			layers of the neural network correspond to $f$, the last to $g$. The
			semi-supervised model corresponds to the \tsvm{} segmentation. We
			optimize on the unlabeled dataset using mean square error (\mse{}) and
			on the labeled dataset using cross-entropy (X-Entropy).}
		\label{fig:ds_diagram}
	\end{center}
\end{figure*}

\subsection{Details: Low-Density Separation}
Upon initializing $f$ and $g$, $f_{\theta_0}$ is a mapping that produces
features well-suited to differentiating between classes. We form $\tilde D_0 =
	\{ (f_{\theta_{0}}(x), y) : (x,y) \in \mathcal{D}_0\}$ and $\tilde D_1 = \{
	f_{\theta_{0}}(x) : x \in \mathcal{D}_1\}$ by passing the feature-data through
this mapping. We then train $c$ \tsvm{}'s, one for each class, on the labeled
data $\tilde D_0$ and unlabeled data $\tilde D_1$.

Our implementation follows Collobert et al.'s \tsvm{}-\cccp{}
method~\cite{Col06} and is based on the R implementation in
\textsc{rssl}~\cite{Kri17}. The algorithm decomposes the \tsvm{} loss function
$J(w,b)$ from \eqref{eq:tsvm_loss} into the sum of a concave function and a
convex function by creating two copies of the unlabeled data, one with positive
labels and one with negative labels. Using the concave-convex procedure
(\textsc{cccp}:~\cite{Yui02,Yui03}), it then reduces the original optimization
problem to an iterative procedure where each step requires solving a convex
optimization problem similar to that of the supervised \svm{}. These convex
problems are then solved using quadratic programming on the dual formulations
(for details, see~\cite{Boy04}). Collobert et al. argue that \tsvm{}-\cccp{}
outperforms previous \tsvm{} algorithms with respect to both speed and
accuracy~\cite{Col06}.

\subsection{Details: Iterative Refinement}
Upon training the \tsvm{}'s, we obtain a probability vector $\hat p_i \in
	\bbbp^c$ for each $i = \ell +1,\dotsc, \ell + u$ with elements corresponding to
the likelihood that $x_i$ lies in a given class. We then form $\breve{\mathcal
		D}_1 = \{(x_i, \hat p_i)\}$ and obtain a supervised training set for further
refining $g \circ f$. We set the learning rate for our Adam optimizer to 1/10th
of its initial rate and minimize the mean square error between $g(f(x_i))$ and
$\hat p_i$ for $(x_i,\hat p_i) \in \breve{\mathcal D}_1$ for 10 epochs
(\emph{cf.} ``consistency loss'' from~\cite{Tar17}) and then minimize the
\textsc{kl}-divergence between $h(y_i)$ and $g(f(x_i))$ for 10 epochs. This
training starts with neural network parameters $\theta_0$ and $\psi_0$ and
produces parameters $\theta_1$ and $\psi_1$. Then, $f_{\theta_1}$ is a mapping
that produces features better suited to segmenting classes than those from
$f_{\theta_0}$. We pass our feature-data through this mapping and continue the
iterative process for $T=6$ iterations. Our settings for learning rate, number
of epochs, and $T$ were hand-chosen for our data and would likely vary for
different applications.

As the algorithm progresses, we store the predictions
$g_{\psi_t}(f_{\theta_t}(x_i))$ at each step $t$ and form an exponential moving
average (discount rate $\rho=0.8$) over them to produce our final estimate for
the probabilities of interest.

\subsection{Remarks on Methodology} \label{sec:hybrid_comparison}
We view our algorithm as most closely related to the work of Iscen et
al.~\cite{Isc19} and Zhuang et al.~\cite{Zhu19}. Both their work and ours
iterate between refining a neural network-based latent representation and
applying a classical \ssl{} method to that representation to produce labels for
further network training. While their work concentrates on graph-based label
propagation, ours uses low-density separation, an approach that we believe may
be more suitable for the task. The representational embedding we learn is
optimized to discriminate between class labels, and for this reason we argue it
makes more sense to refine decision boundaries than it does to pass labels.
Additionally, previous work on neural network-based classification suggests
that an \svm{} loss function can improve classification accuracy~\cite{Tan13},
and our data augmentation step effectively imposes such a loss function for
further network training.

By re-learning decision boundaries at each iterative step, we allow our
algorithm to recover from mistakes it makes in early iterations. One failure
mode of semi-supervised methods entails making a few false label assignments
early in the iterative process and then having these mislabeled points pass
these incorrect labels to their neighbors. For example, in
pseudo-labelling~\cite{Lee13}, the algorithm augments the underlying training
set $\mathcal{D}_0$ with pairs $(x_i, \hat y_i)$ for $x_i \in \mathcal{D}_1$
and predicted labels $\hat y_i$ for which the model was most confident in the
previous iteration. Similar error-reinforcement problems can occur with
boosting~\cite{Mal09}. It is easy to see how a few confident, but inaccurate,
labels that occur in the first few steps of the algorithm can set the labeling
process completely askew.

By creating an embedding $f:\bbbr^d\to\bbbr^m$ and applying linear separation
to embedded points, we have effectively learned a distance metric $\kappa:
	\bbbr^d\times\bbbr^d\to\bbbr_{\ge 0}$ especially suited to our learning
problem. The linear decision boundaries we produce in $\bbbr^m$ correspond to
nonlinear boundaries for our original features in $\bbbr^d$. Previously, Jean
et al. \cite{Jea18} described using a deep neural network to embed features for
Gaussian process regression, though they use a probabilistic framework for
\ssl{} and consider a completely different objective function.

\section{Application to User Classification from Survey Data}

In this section, we discuss the practical problem of segmenting users from
survey data and compare the performance of our algorithm to other
recently-developed methods for \ssl{} on real data. We also perform an ablation
study to ensure each component of our process contributes to the overall
effectiveness of the algorithm.

\subsection{Description of the Dataset}
At Adobe, we are interested in segmenting users based on their work habits,
artistic motivations, and relationship with creative software. To gather data,
we administered detailed surveys to a select group of users in the US, UK,
Germany, \& Japan (just over 22 thousand of our millions of users). We applied
Latent Dirichlet Allocation (\textsc{lda}:~\cite{Pri00,Ble03}), an unsupervised
model to discover latent topics, to one-hot encoded features generated from
this survey data to classify each surveyed user as belonging to one of $c=5$
different segments. We generated profile and usage features using an in-house
feature generation pipeline (that could in the future readily be used to
generate features for the whole population of users). In order to be able to
evaluate model performance, we masked the \textsc{lda} labels from our surveyed
users at random to form the labelled and unlabelled training sets
$\mathcal{D}_0$ and $\mathcal{D}_1$.

\subsection{State-of-the-Art Alternatives}

We compare our algorithm against two popular classification algorithms. We
focus our efforts on other algorithms we might have actually used in practice
instead of more similar methods that just recently appeared in the literature.

The first, LightGBM~\cite{Ke17} is a supervised method that attempts to improve
upon other boosted random forest algorithms (e.g. the popular
xgBoost~\cite{Che16}) using novel approaches to sampling and feature bundling.
It is our team's preferred nonlinear classifier, due to its low requirements
for hyperparameter tuning and effectiveness on a wide variety of data types. As
part of the experiment, we wanted to evaluate the conditions for
semi-supervised learning to outperform supervised learning.

The second, Mean Teacher~\cite{Tar17} is a semi-supervised method that creates
two supervised neural networks, a teacher network and a student network, and
trains both networks using randomly perturbed data. Training enforces a
consistency loss between the outputs (predicted probabilities in $\bbbp^c$) of
the two networks: optimization updates parameters for the student network and
an exponential moving averages of these parameters become the parameters for
the teacher network. The method builds upon Temporal Ensembling~\cite{Lai17}
and uses consistency loss~\cite{Ras15,Saj16}.

\subsection{Experimental Setup}
We test our method with labelled training sets of successively increasing size
$\ell \in \{35, 50, 125, 250, 500, 1250, 2500\}$. Each training set is a strict
superset of the smaller training sets, so with each larger set, we strictly
increase the amount of information available to the classifiers. To tune
hyperparameters, we use a validation set of size 100, and for testing we use a
test set of size 4780. The training, validation, and test sets are selected to
all have equal class sizes.

For our algorithm, we perform $T=6$ iterations of refinement, and in the
\tsvm{} we set the cost parameters $C = 0.1$ and $C^* = \frac{\ell}{u} C$. To
reduce training time, we subsample the unlabeled data in the test set by
choosing 250 unlabeled points uniformly at random to include in the \tsvm{}
training. We test using our own implementations of \tsvm{} and MeanTeacher.

\subsection{Numerical Results and Ablation}

\begin{figure}[t]
	\begin{center}
		\includegraphics[width=\linewidth]{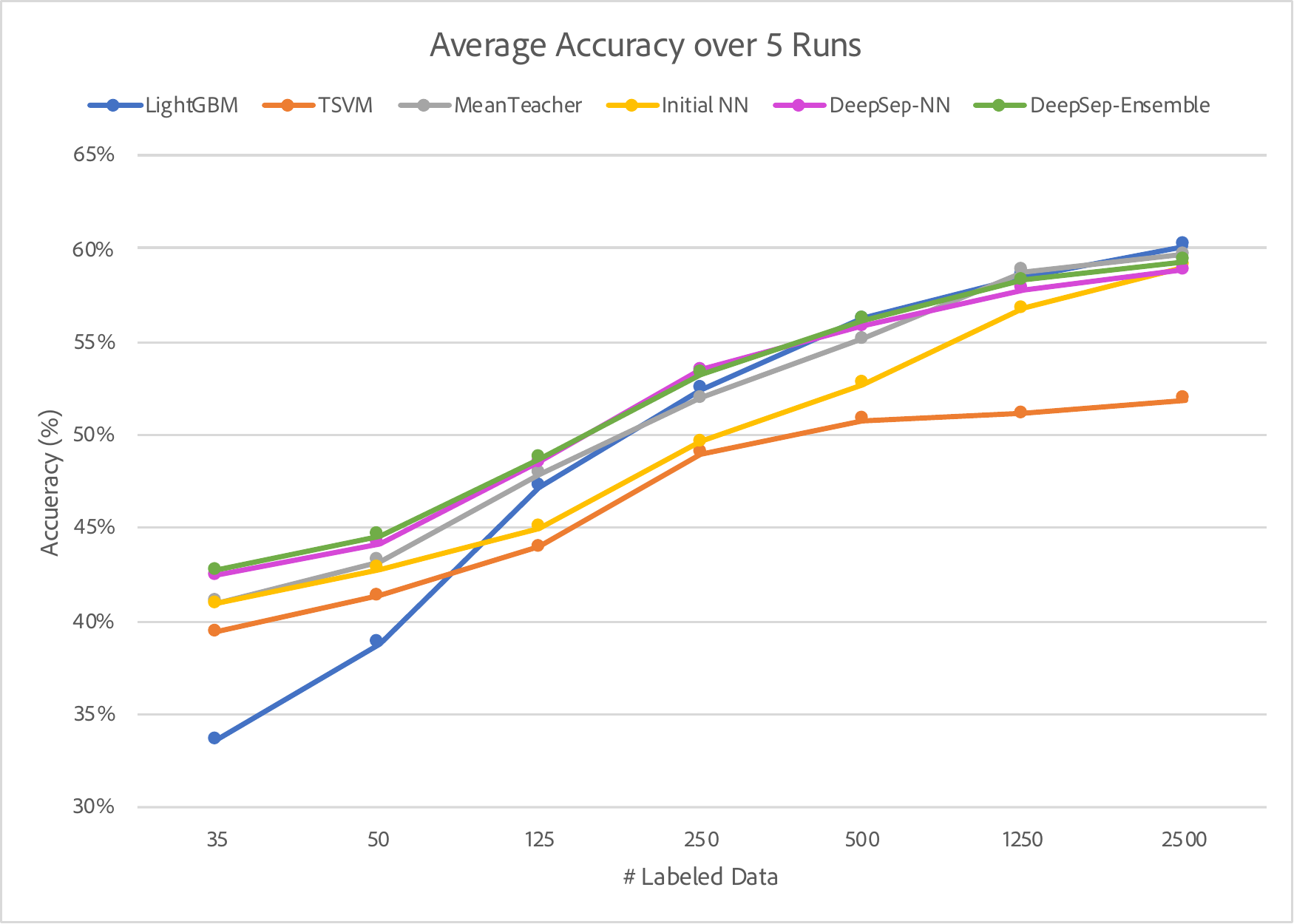}
		\caption{Average accuracy over 5 random shuffles for LightGBM, \tsvm{},
			MeanTeacher and our proposed method. Random chance accuracy is 20\%. We
			are primarily interested in the regime where few training examples
			exist -- particularly when the number of labeled datapoints is 35-50.}
		\label{fig:avgacc}
	\end{center}
\end{figure}

Table~\ref{table:results} reports our classification accuracy on five
randomized shuffles of the training, validation, and test sets. These results
are summarized in Figure~\ref{fig:avgacc}. The accuracy of our baseline methods
are shown first, followed by three components of our model:
\begin{enumerate}
	\item Initial NN: The output of the neural network after initial supervised
	training.
	\item DeepSep-NN: The output of the neural network after iterative
	refinement with Algorithm~\ref{algo}.
	\item DeepSep-Ensemble: Exponential moving average as described in
	Algorithm~\ref{algo}.
\end{enumerate}

We find that Deep Low-Density Separation outperforms or matches LightGBM in the
range $\ell \leq 1250$. The relative increase in accuracy of Deep Separation is
as much as 27\%, which is most pronounced with a very small amount of training
data ($\ell \leq 50$). Some of this increase can be attributed to the initial
accuracy of the neural network; however, the iterative refinement of Deep
Separation improves the accuracy of the initial network by up to 8.3\%
(relative). The addition of a temporal ensemble decreases variance in the final
model, further increasing accuracy by an average of 0.54\% across the range.
Compared to Mean Teacher, the iterative refinement of Deep Separation achieves
a larger increase in accuracy for $l \leq 500$.

\begin{table}[t]
	\centering
	\caption{Classification accuracy (in percent) for each of the methods
		tested. Shuffle \# refers to the randomized splitting of the data into
		training, validation, and test sets. The final block contains the average
		accuracy over 5 random shuffles.}
	\label{table:results}
	\begin{tabular}{|l|l|c c c c c c c|}
		\hline
		           &                  & $\ell$ &       &       &       &       &       &       \\
		Shuffle \# & Model            & 35     & 50    & 125   & 250   & 500   & 1250  & 2500  \\
		\hline \hline

		1          & LightGBM         & 30.98  & 34.73 & 47.45 & 51.55 & 55.99 & 59.39 & 60.65 \\
		           & \tsvm{}          & 38.65  & 38.26 & 40.26 & 46.84 & 48.54 & 51.02 & 52.94 \\
		           & MeanTeacher      & 39.91  & 41.70 & 47.54 & 51.33 & 54.81 & 59.83 & 60.48 \\
		           & Initial NN       & 38.65  & 40.09 & 41.92 & 47.89 & 51.15 & 58.13 & 61.09 \\
		           & DeepSep-NN       & 39.04  & 41.79 & 44.97 & 53.55 & 54.51 & 57.60 & 60.13 \\
		           & DeepSep-Ensemble & 40.13  & 42.00 & 46.32 & 52.68 & 54.95 & 58.04 & 59.87 \\\hline
		2          & LightGBM         & 32.03  & 38.30 & 46.93 & 52.33 & 55.86 & 58.39 & 59.61 \\
		           & \tsvm{}          & 43.31  & 43.88 & 47.45 & 49.19 & 50.76 & 49.76 & 50.37 \\
		           & MeanTeacher      & 43.14  & 42.75 & 48.58 & 53.03 & 54.12 & 58.08 & 59.35 \\
		           & Initial NN       & 43.31  & 43.79 & 45.97 & 50.72 & 53.03 & 57.04 & 59.26 \\
		           & DeepSep-NN       & 47.32  & 47.10 & 48.85 & 51.90 & 54.25 & 56.69 & 57.95 \\
		           & DeepSep-Ensemble & 46.45  & 46.58 & 49.06 & 51.94 & 54.47 & 57.60 & 58.56 \\\hline
		3          & LightGBM         & 32.33  & 40.31 & 47.63 & 50.94 & 56.34 & 57.82 & 60.13 \\
		           & \tsvm{}          & 30.37  & 34.55 & 37.30 & 49.93 & 51.59 & 52.42 & 51.42 \\
		           & MeanTeacher      & 35.77  & 40.26 & 45.05 & 50.33 & 55.12 & 56.43 & 57.25 \\
		           & Initial NN       & 37.12  & 40.87 & 43.05 & 48.15 & 52.72 & 55.82 & 57.86 \\
		           & DeepSep-NN       & 36.69  & 40.48 & 46.88 & 52.33 & 55.60 & 56.95 & 57.82 \\
		           & DeepSep-Ensemble & 37.17  & 40.52 & 46.49 & 52.33 & 56.12 & 57.04 & 57.86 \\\hline
		4          & LightGBM         & 35.12  & 36.17 & 47.36 & 52.42 & 56.30 & 59.00 & 61.05 \\
		           & \tsvm{}          & 40.61  & 45.10 & 48.28 & 52.85 & 52.64 & 50.11 & 51.29 \\
		           & MeanTeacher      & 41.96  & 44.31 & 49.54 & 51.76 & 55.56 & 59.56 & 60.96 \\
		           & Initial NN       & 41.26  & 43.66 & 48.10 & 48.63 & 52.55 & 55.64 & 58.26 \\
		           & DeepSep-NN       & 44.84  & 44.58 & 50.41 & 54.34 & 56.86 & 58.61 & 59.08 \\
		           & DeepSep-Ensemble & 44.49  & 44.88 & 50.33 & 53.46 & 56.86 & 59.39 & 60.44 \\\hline
		5          & LightGBM         & 37.60  & 44.44 & 46.67 & 55.16 & 56.60 & 57.95 & 59.30 \\
		           & \tsvm{}          & 44.14  & 45.14 & 46.71 & 46.06 & 50.41 & 52.24 & 53.51 \\
		           & MeanTeacher      & 44.14  & 46.93 & 48.63 & 53.25 & 56.08 & 60.17 & 60.22 \\
		           & Initial NN       & 44.44  & 45.62 & 45.88 & 52.85 & 54.29 & 57.39 & 58.69 \\
		           & DeepSep-NN       & 44.44  & 46.93 & 51.46 & 55.38 & 58.00 & 59.39 & 59.17 \\
		           & DeepSep-Ensemble & 45.53  & 48.85 & 51.37 & 55.90 & 58.43 & 59.48 & 59.96 \\\hline \hline
		Average    & LightGBM         & 33.61  & 38.79 & 47.21 & 52.48 & 56.22 & 58.51 & 60.15 \\
		           & \tsvm{}          & 39.42  & 41.39 & 44.00 & 48.98 & 50.79 & 51.11 & 51.90 \\
		           & MeanTeacher      & 40.98  & 43.19 & 47.87 & 51.94 & 55.14 & 58.81 & 59.65 \\
		           & Initial NN       & 40.96  & 42.81 & 44.98 & 49.65 & 52.75 & 56.80 & 59.03 \\
		           & DeepSep-NN       & 42.47  & 44.17 & 48.51 & 53.50 & 55.84 & 57.85 & 58.83 \\
		           & DeepSep-Ensemble & 42.75  & 44.57 & 48.71 & 53.26 & 56.17 & 58.31 & 59.34 \\
		\hline
	\end{tabular}
\end{table}

\begin{figure}[t]
	\begin{center}
		\includegraphics[width=\linewidth]{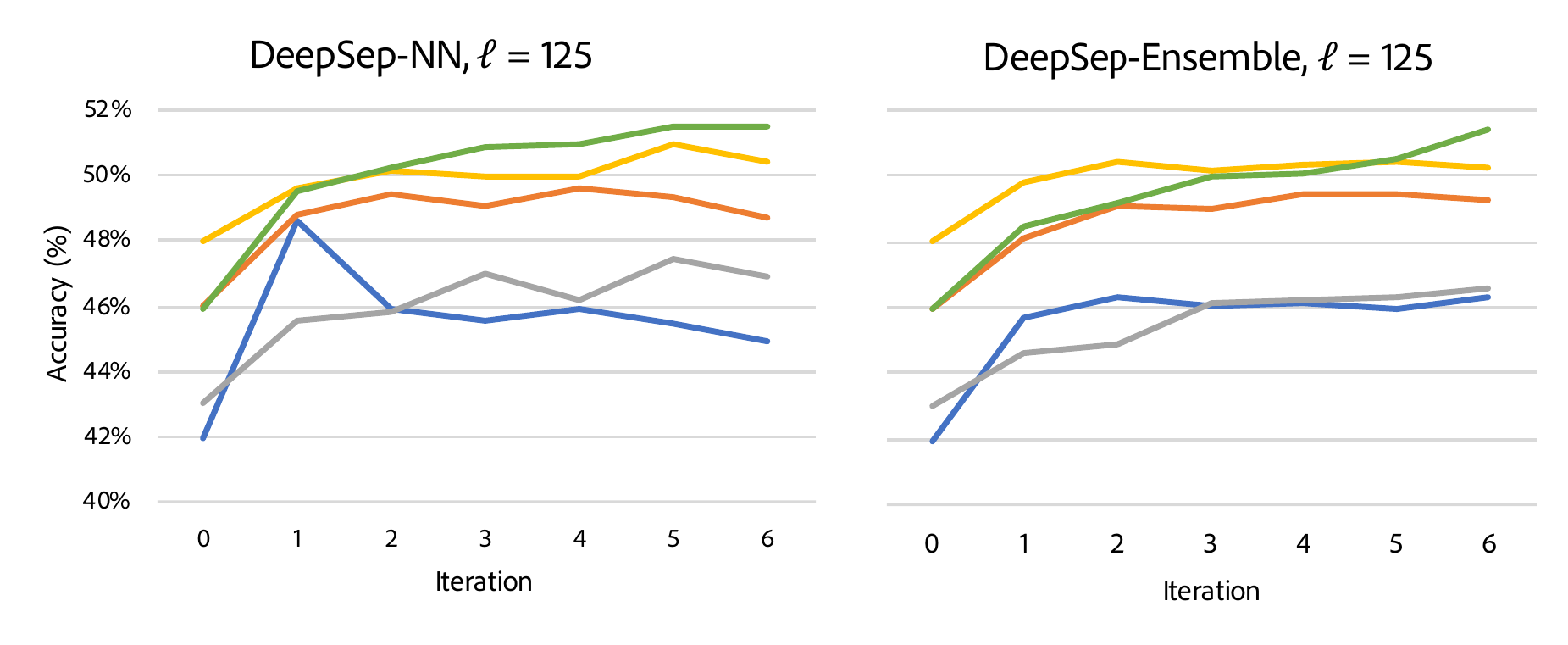}
		\caption{Classification accuracy on the test set for all five random
			shuffles over the course of iterative refinement, using 125 labeled
			data, of (left) the refined neural network and (right) the exponential
			moving average of predictions. Here, different colors correspond to
			different choices for training set (different random seeds).}
		\label{fig:iterations}
	\end{center}
\end{figure}

To visualize how the iterative refinement process and exponential weighted
average improve the model, Figure \ref{fig:iterations} shows the accuracy of
our model at each iteration. We see that for each random shuffle, the
refinement process leads to increased accuracy compared to the initial model.
However, the accuracy of the neural network fluctuates by a few percent at a
time. Applying the exponential moving average greatly reduces the impact of
these fluctuations and yields more consistent improvement, with a net increase
in accuracy on average.

Regarding training time, all numerical experiments were performed on a mid-2018
MacBook Pro (2.6 GHz Intel Core i7 Processor; 16 GB 2400 MHz DDR4 Memory). Deep
Separation takes up to half an hour on the largest training set ($\ell =
	2500$). However, we note that for $\ell \leq 500$, the model takes at most
three minutes, and this is the regime where our method performs best in
comparison to other methods. In contrast, LightGBM takes under a minute to run
with all training set sizes.

\section{Conclusions}
In this paper, we introduce a novel hybrid semi-supervised learning method,
Deep Low-Density Separation, that iteratively refines a latent feature
representation and then applies low-density separation to this representation
to augment the training set. We validate our method on a multi-segment
classification dataset generated from surveying Adobe's user base. In the
future, we hope to further investigate the interplay between learned feature
embeddings and low-density separation methods, and experiment with different
approaches for both representational learning and low-density separation. While
much of the recent work in deep \ssl{} concerns computer vision problems and
image classification in particular, we believe these methods will find wider
applicability within academia and industry, and anticipate future advances in
the subject.


\end{document}